%% file: template.tex
  \RequirePackage{fix-cm}
  \documentclass[natbib,twocolumn]{svjour3}          
  \smartqed  
\usepackage{pifont}
\usepackage{amsmath,amsfonts}
\usepackage{algorithmic}
\usepackage{algorithm}
\usepackage{array}
\usepackage{microtype}
\usepackage{xspace}
\usepackage{esvect}
\usepackage[caption=false,font=normalsize,labelfont=sf,textfont=sf]{subfig}
\usepackage{textcomp}
\usepackage{stfloats}
\usepackage{url}
\usepackage{verbatim}
\usepackage{graphicx}
\usepackage{multirow}
\newcolumntype{Y}{>{\centering\arraybackslash}X}
\newcolumntype{H}{>{\setbox0=\hbox\bgroup}c<{\egroup}@{}}
\usepackage{graphicx}
\usepackage{amsmath}
\usepackage{amssymb}
\usepackage{booktabs}
\usepackage{mathrsfs}
\usepackage{mathtools}
\usepackage[table]{xcolor}
\usepackage[pagebackref=true,breaklinks=true,citecolor=tealblue,colorlinks,linkcolor=ruby,bookmarks=false]{hyperref}
\definecolor{ruby}{rgb}{0.88, 0.07, 0.37}
\definecolor{tealblue}{rgb}{0.18, 0.40, 0.46}

\begin{document}

\title{3DDesigner: Towards Photorealistic 3D Object Generation and Editing with Text-guided Diffusion Models}

\author{Gang Li$^{1,2}$ \and Heliang Zheng$^{3}$ \and Chaoyue Wang$^{3}$ \and Chang Li$^{3}$ \and \\ Changwen Zheng $^{1}$ \and Dacheng Tao$^{3}$}

\institute{
  \@institute {\ Changwen Zheng (\textit{Corresponding Author.})} \at \email{changwen@iscas.ac.cn} \and
  $^1$ Institute of Software, Chinese Academy of Sciences, Beijing, China. \\
  $^2$ University of Chinese Academy of Sciences, Beijing, China.\\
  $^3$ JD Explore Academy, Beijing, China.\\
}

\date{Received: October 7, 2023 / Accepted: xxx }

\def\smallgap{\vspace{0.05in}}
\maketitle
\begin{abstract}
Text-guided diffusion models have shown superior performance in image/video generation and editing. While few explorations have been performed in 3D scenarios. In this paper, we discuss three fundamental and interesting problems on this topic. First, we equip text-guided diffusion models to achieve \textbf{3D-consistent generation}. Specifically, we integrate a NeRF-like neural field to generate low-resolution coarse results for a given camera view. Such results can provide 3D priors as condition information for the following diffusion process. During denoising diffusion, we further enhance the 3D consistency by modeling cross-view correspondences with a novel two-stream (corresponding to two different views) asynchronous diffusion process. Second, we study \textbf{3D local editing} and propose a two-step solution that can generate 360$^{\circ}$ manipulated results by editing an object from a single view. Step 1, we propose to perform 2D local editing by blending the predicted noises. Step 2, we conduct a noise-to-text inversion process that maps 2D blended noises into the view-independent text embedding space. Once the corresponding text embedding is obtained, 360$^{\circ}$ images can be generated. Last but not least, we extend our model to perform \textbf{one-shot novel view synthesis} by fine-tuning on a single image, firstly showing the potential of leveraging text guidance for novel view synthesis. Extensive experiments and various applications show the prowess of our 3DDesigner. The project page is available at {\url{https://3ddesigner-diffusion.github.io/}}
\end{abstract}

\begin{figure*}[!t]
    \centering
    \includegraphics[width=1\linewidth]{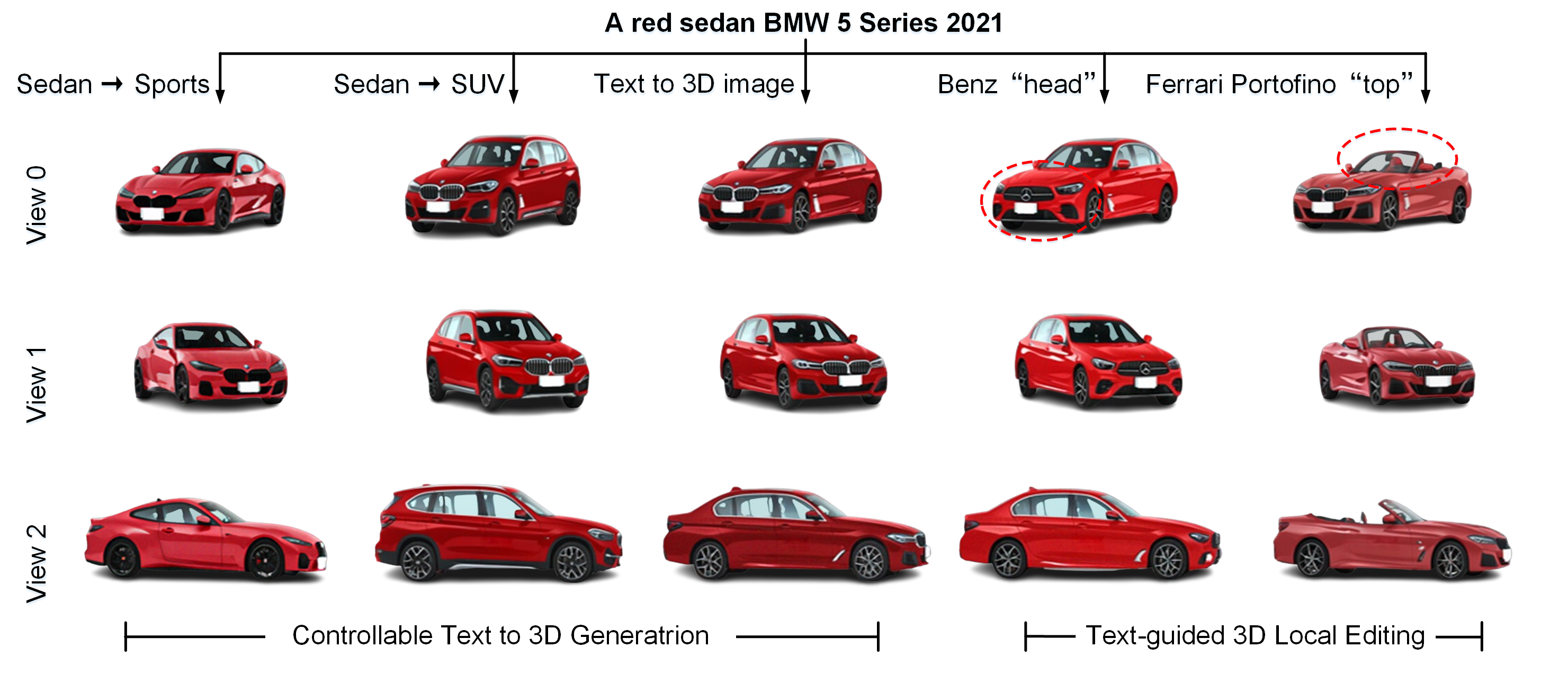}
    \caption{``3DDisigner'' -- Text guided 3D object generation and editing. Given a text, \textit{e.g.,} ``A red sedan BMW 5 series 2021'', our method can 1) generate the corresponding 3D images, 2) create ``SUV'' and ``Sports'' 3D counterparts, and 3) support text-guided 3D local editing.}
    \label{fig:teaser}
\end{figure*}

\input{latex/2introduction}

\input{latex/3relatedwork}

\input{latex/4method}
\input{latex/5experiment}

\input{latex/6conclusion}

\bibliographystyle{spbasic}      
\bibliography{egbib}

\end{document}

%% file: latex/2introduction.tex
\section{Introduction}
\label{sec:intro}
Recently, the advent of text-guided diffusion models has brought about a paradigm shift in the field of image synthesis, offering remarkable advancements in the creation of high-fidelity, diverse, and highly controllable images~\cite{Glide, DALLE2, Imagen}. These models have sparked the rise of what can be termed ``AI artists," capable of producing intricate and imaginative artworks within mere seconds. Moreover, substantial strides have been achieved in the domain of image editing~\cite{kim2022diffusionclip, avrahami2022blended,lugmayr2022repaint,gal2022image, Dreambooth,kawar2022imagic}, enabling complex and nuanced text-guided semantic modifications, even for objects with non-rigid structures, using nothing more than a simple text prompt. Beyond static images, text-guided diffusion models have been successfully extended to the realm of video synthesis~\cite{vdm, Make-A-Video, imagenvideo, villegas2022phenaki}, achieving remarkable results across multiple dimensions, including enhanced data efficiency, the production of high-definition video content, and the extension of video length. 

Real-world objects are 3D, and it usually requires multi-view information to represent every detail of an object. Creating 3D assets demands specialized software and skills, and the process can be time-consuming compared to generating 2D content.  Consequently, a tool capable of 3D object generation holds great significance for real-world applications.  However, limited exploration has been conducted on text-guided diffusion models in 3D scenarios.  CLIP-NeRF~\cite{wang2022clip} integrate a CLIP~\cite{CLIP} model to guide the rendering of a Neural Radiance Field (NeRF)~\cite{NeRF}, while without the generation capability of diffusion models, CLIP-NeRF cannot achieve flexible control and local editing. 3DiM~\cite{3ddiffusion} firstly makes a diffusion model work well for 3D novel view synthesis, while without text guidance, it is hard to perform novel object generation and manipulation. Dreamfusion~\cite{Dreamfusion} creatively adopt text-guided diffusion models as supervision to train a NeRF, but such a solution is not efficient, \textit{i.e.,} an optimized model can only generate one specific object.

In this paper, we delve into text-guided diffusion models for 3D object generation and manipulation (Figure~\ref{fig:teaser}). We originally discuss three fundamental and interesting problems. The quintessential and arguably the most crucial concern: the attainment of \textbf{3D-consistent generation}. Our proposed model comprising a NeRF-based condition module and a two-stream asynchronous diffusion module. The NeRF-based condition module takes camera views and coarse text guidance (\textit{e.g.,} ``A red sedan BMW'') and generates low-resolution coarse results for the given camera views as inputs. Such results can provide 3D priors, serving as condition information for the following diffusion process. The two-stream asynchronous diffusion module is designed to further enhance the 3D consistency by modeling cross-view correspondences. This approach entails the joint denoising of two distinctively noised images, each derived from varying perspectives (one perspective per stream) of same object, where the cross-view feature interactions can encourage the two streams to generate images that are consistent with each other. During sampling, one of the two streams loads previously generated views to guide the generation of the current view. In this way, 360$^{\circ}$ consistent results can be obtained.

Secondly, we explore \textbf{3D local editing}. To the best of our knowledge, we are the first to perform 360$^{\circ}$ manipulated results by editing an object from a single view. We achieve this in two steps. In Step 1, we perform 2D local editing. We propose a ``noise blending'' pipeline to edit a specific region of a generated image. Specifically, during each sampling step, we predict the noise under the guidance of the provided text and replace a defined region of this noise with the noise predicted by the target text.  This allows us, for example, to seamlessly combine the body of a BMW Series 5 car with the front end of a Benz Class-E car. In Step 2, we design a noise-to-text inversion process that maps 2D blended noises into the view-independent text embedding space. Once the corresponding text embedding is obtained, 360$^{\circ}$ images can be generated. 

Last but not least, we discover that our model can also be easily extended to perform \textbf{one-shot novel view synthesis} by simply fine-tuning on a single image. Impressive results demonstrate the potential of leveraging text guidance for novel view synthesis, and we hope our work provides valuable insights for the novel view synthesis community.

%% file: latex/3relatedwork.tex
\section{Related Work}
\label{sec:relate}

\subsection{Text-guided diffusion models}
Diffusion models~\cite{sohl2015deep,song2019generative,ddpm} have made tremendous progress recently, showing superiority on stable training, high-fidelity image/video synthesis, non-trivial semantic editing, multi-modality fusion, etc ~\cite{dhariwal2021diffusion, Glide, stable,avrahami2022blended,gal2022image,kawar2022imagic, Make-A-Video,villegas2022phenaki, DALLE2}. Meanwhile, billions of (text, image) pairs~\cite{schuhmann2022laion} enable the recent breakthroughs in text-to-image synthesis~\cite{ramesh2021zero,DALLE2,Imagen, stable,yu2022scaling}. Thus not surprisingly, text-guided diffusion models have shown impressive performance and attracted lots of attention. In this work, we further explore the performance of text-guided diffusion models for 3D generation. 

Once a text-guided diffusion model is trained, it can be further extended to perform image editing. GLIDE ~\cite{Glide} formulate an in-painting task, where the text can guide the generation of the masked region. Blend Diffusion~\cite{avrahami2022blended} and Repaint~\cite{lugmayr2022repaint} spatially blend noised versions of the input image with the local text-guided diffusion latent at a progression of noise levels. Textual inversion~\cite{gal2022image} and DreamBooth~\cite{Dreambooth} propose the concept of ``personalizing generation'', which inverts an object to a special text token.  Imagic~\cite{kawar2022imagic} propose a fine-tuning technique that enables various text-based semantic edits on a single real input image. In this work, we extend noisy image blending to noise blending and further conduct noise-to-text inversion for 3D local editing.

\begin{figure*}[!t]
    \centering
    \includegraphics[width=1\linewidth]{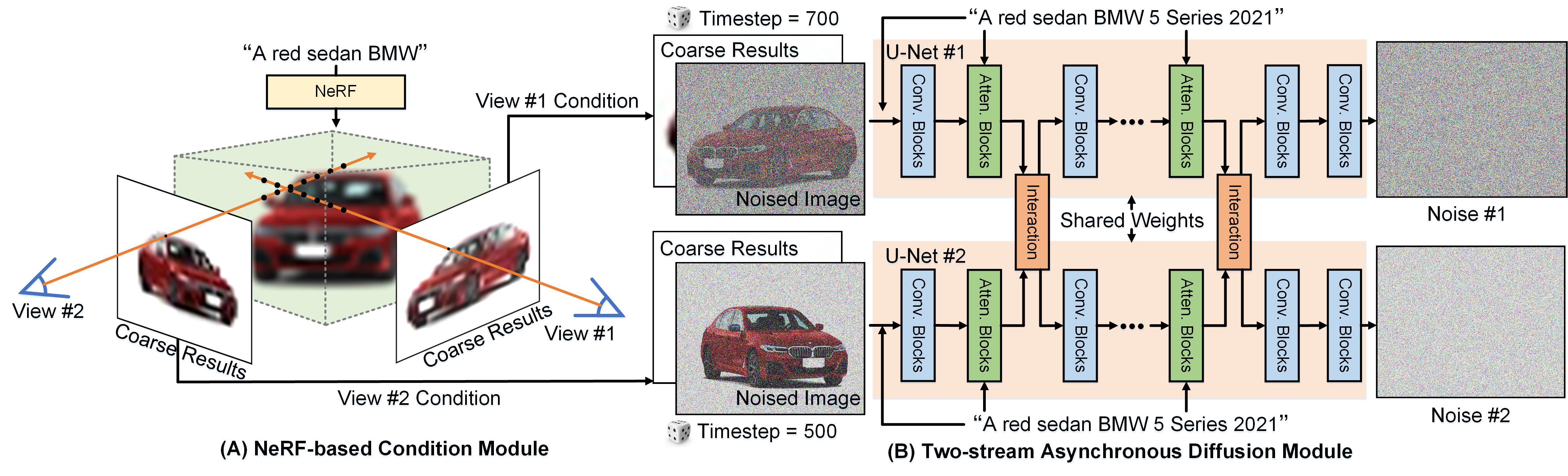}
    \caption{An illustration of our framework for text-guided 3D-consistent generation (training phase). (A) NeRF-based Condition Module, which takes $<$one coarse text, two camera views$>$ pairs as inputs and generates low-resolution coarse results. The coarse results are resized and concatenated with noised images to provide conditions for denoising. (B) Two-stream Asynchronous Diffusion Module, which takes $<$one full text, two coarse results, two timesteps, two noised images$>$ quadruples as inputs and predicts the added noises. Each stream is a vanilla text-guided diffusion model except for the feature interaction module after each attention block. Note that the timesteps are randomly generated and the parameters of these two streams are shared.}
    \label{fig:framework}
\end{figure*}

\subsection{3D novel view synthesis}
Since NeRF~\cite{NeRF} was firstly proposed to learn implicit 3D representations for the task of novel view synthesis, plenty of variants have been proposed to improve the efficiency~\cite{zhang2021nerfactor,mueller2022instant,zhu2023pyramid}, generalization~\cite{chen2021mvsnerf,johari2022geonerf,yu2021pixelnerf,sharma2022seeing,rebain2022lolnerf, lan2023correspondence}, controllability~\cite{yang2021learning,wang2022clip,lazova2022control}, etc. For example, CLIP-NeRF~\cite{wang2022clip} integrates CLIP~\cite{CLIP} with NeRF and makes the colors and shapes can be controlled by text. Some recent literature also proposes to get rid of explicit geometric inductive biases (\textit{e.g.,} introduced by volume rendering). 3DiM~\cite{3ddiffusion} firstly makes a diffusion model work well for 3D novel view synthesis. Moreover, Dreamfusion~\cite{Dreamfusion} creatively adopt text-guided diffusion models as supervision to train a NeRF. Although GAUDI~\cite{bautista2022gaudi} also integrates NeRF with diffusion models, the diffusion model here is only used to learn a controllable latent code, preventing their model from performing local editing. Compared to these works, we leverage text-guided diffusion models to achieve more controllable and editable 3D generation.

%% file: latex/4method.tex
\section{Approach}
\label{sec:method}

In this section, we introduce the details of our proposed 3DDesigner, including 1) 3D-consistent generation, \textit{i.e.,} learning a text-guided diffusion model that can generate images from different camera views with consistent content; 2) 3D local editing, \textit{i.e.,} generating 360$^{\circ}$ manipulated results by editing an object from a single view; and 3) one-shot novel view synthesis, \textit{i.e.,} synthesising 360$^{\circ}$ results based on a single image.

\subsection{3D-consistent generation}

Figure~\ref{fig:framework} is an illustration of our framework for text-guided 3D-consistent generation (training phase). Specifically, we integrate a NeRF-like neural field to generate 3D-consistent coarse results, which are leveraged in the following denoising diffusion process in the form of image conditions. For the denoising diffusion process, we propose to jointly denoise two noised images of an object with different views. So that the cross-view feature interactions can encourage the two streams to generate images that are consistent with each other. For now, the two views' consistency can be achieved. To further obtain 360$^{\circ}$ consistent results, we adopt an autoregressive sampling strategy~\cite{3ddiffusion}, where one stream loads previously generated views to guide the other stream to generate the current view.

\noindent \textbf{NeRF-based condition module.}
The NeRF-based condition module takes $<$coarse text, camera view$>$ pairs as inputs and generates coarse results. Note that we do not use the full-text guidance here to avoid suppressing the creative ability of the following denoising diffusion module (\textit{i.e.,} making it a simple super-resolution module). 

\underline{\textit{Learning MLP.}} We first calculate the 3D location and view direction of each camera ray according to the given camera view. The camera ray can be denoted as $\mathbf{r}(t) = \mathbf{o} + t\mathbf{d}$, where $\mathbf{o}$ and $\mathbf{d}$ are the origin and direction of the ray, respectively. We use a multiple layer perceptual (MLP) to predict the density $\sigma(t)$ and RGB color $\mathbf{c}(t)$ of a given point in the NeRF implicit representation:
\begin{equation}\label{eqn:nerf}
\sigma(t), \mathbf{c}(t) = MLP(\mathbf{r}(t),\mathbf{d},\mathbf{y}_c) 
\end{equation}
where $\mathbf{y}_c$ denotes the coarse text embedding. In particular, we concatenate the position embedded of $\mathbf{r}(t)$ and the text embedding $\mathbf{y}_c$ before getting through the MLP.

\underline{\textit{Volume rendering.}} We can obtain coarse results by conducting volume rendering:
\begin{equation}\label{eqn:render}
\mathbf{x}_c(\mathbf{r}) = \int_{t_n}^{t_f} T(t)\sigma(t)\mathbf{c}(t) dt 
\end{equation}
where $t_n$ and $t_f$ are the near and far bounds, respectively. $T(t) = exp(- \int_{t_n}^{t} \sigma(s) ds)$ handles occlusion, and $\mathbf{x}_c(\mathbf{r})$ indicates the RGB value of the coarse result $\mathbf{x}_c$ that rendered by camera ray $\mathbf{r}$. In our experiments, we follow StyleNeRF~\cite{gu2021stylenerf}, 1) approximating the volume rendering process to a GPU-friendly version that supports operations on a set of rays and 2) removing the view direction condition to suppress spurious correlations and dataset bias.

\noindent \textbf{Two-stream asynchronous diffusion module.} The two-stream asynchronous diffusion module takes $<$full text, coarse results, timesteps, noised images$>$ quadruples as inputs and predicts the added noises.

\underline{\textit{Diffusion.}} Given two images of an object sampled from different views, $\mathbf{x}^1$ and $\mathbf{x}^2$, the diffusion processes (\textit{i.e.,} adding noise) for these two images are independent, which can be denoted as:
\begin{equation}\label{eqn:add_noise}
\begin{split}
  & q(\mathbf{x}^1_t|\mathbf{x}^1_{0}) \coloneqq \mathcal{N}(\mathbf{x}^1_t;\sqrt{\bar\alpha^1_t}\mathbf{x}^1_{0},(1-\bar\alpha^1_t)\mathbf{I}), \\
& q(\mathbf{x}^2_t|\mathbf{x}^2_{0}) \coloneqq \mathcal{N}(\mathbf{x}^2_t;\sqrt{\bar\alpha^2_t}\mathbf{x}^2_{0},(1-\bar\alpha^2_t)\mathbf{I}),  
\end{split}
\end{equation}

\noindent where $t$ is the timestep, $\bar\alpha_t$ can be calculated by the variance schedule, and $\mathbf{I}$ is an identity matrix. 

\underline{\textit{Posterior.}} Now let's move on to approximate the posterior (\textit{i.e.,} reverse process), where all of the coarse results, cross-view information, and text guidance can be leveraged: 
\begin{equation}\label{eqn:denoise}
\begin{split}
& p_\theta(\mathbf{x}^1_{t_1-1}| \mathbf{x}^1_{t_1},\mathbf{x}^1_c,\mathbf{x}^2_{t_2},\mathbf{x}^2_c,\mathbf{y}) \coloneqq \\ &\mathcal{N} (\mu_\theta(\mathbf{x}^1_{t_1},\mathbf{x}^1_c,\mathbf{x}^2_{t_2},\mathbf{x}^2_c,\mathbf{y}), \Sigma_\theta(\mathbf{x}^1_{t_1},\mathbf{x}^1_c,\mathbf{x}^2_{t_2},\mathbf{x}^2_c,\mathbf{y})),
\end{split}
\end{equation}
where $\mathbf{y}$ is the full text, $\mathbf{x}_c$ is the coarse result obtained by Equation~\ref{eqn:render}, $t_1$ and $t_2$ are randomly generated timesteps, $\mu_\theta$ and $\Sigma_\theta$ are models that predict the mean and variance of the Gaussian distribution, respectively. 

\underline{\textit{Optimization.}} In practice, we follow the widely used variational lower-bound re-weighing to learn the posterior. Specifically, we generate samples $\{\mathbf{x}^1_{t_1},\mathbf{x}^2_{t_2}\}$ by applying Gaussian noise $\epsilon \sim \mathcal{N}(\mathbf{0},\mathbf{I})$ to $\{\mathbf{x}^1_{0},\mathbf{x}^2_{0}\}$ using Equation~\ref{eqn:add_noise}. Then we train a model $\epsilon_\theta$ to predict the added noise:
\begin{equation}\label{eqn:loss}
L = \mathbb{E}_{\mathbf{y},\mathbf{x}^1_{0},\mathbf{x}^2_{0},t_1,t_2,\epsilon} 	\parallel \epsilon_\theta(\mathbf{x}^1_{t_1},\mathbf{x}^1_c,\mathbf{x}^2_{t_2},\mathbf{x}^2_c,\mathbf{y}) - \epsilon	\parallel.
\end{equation}

\underline{\textit{Architecture.}} 
We follow the design of multi-stream U-Nets that are widely obtained in video generation~\cite{vdm, Make-A-Video} and 3DiM~\cite{3ddiffusion}. Specifically, the two-stream asynchronous diffusion module consists of two U-Nets and several feature interaction modules. The interaction module can be the temporal convolution/attention used in Make-A-Video~\cite{Make-A-Video} or the cross-attention used in 3DiM~\cite{3ddiffusion} (the latter performs better on our task). The coarse results obtained by Equation~\ref{eqn:render} are resized and concatenated with noised images obtained by Equation~\ref{eqn:add_noise}. The text embedding is integrated by attention blocks, and the timestep is specified by adding position embedding into each convolution block~\cite{ddpm}. 

\underline{\textit{Sampling.}} 
The cross-view feature interaction can encourage the two streams to generate images that are consistent with each other in the sampling phase. Note that we adopt different timesteps to generate different-level noises. Our insight is to make the ``cleaner'' (condition) stream to guide the generation of the ``noisier'' (generation) stream. In particular, we adopt the DDIM~\cite{ddim} sampling process:
\begin{equation}\label{eqn:sampling}
\mathbf{x}^1_{t_1-1} = \sqrt{\alpha^1_{t_1-1}}(\frac{\mathbf{x}^1_{t_1} - \sqrt{1-\alpha^1_{t_1}}\hat{\epsilon}_{t_1}}{\sqrt{\alpha^1_{t_1}}})+\sqrt{1-\alpha^1_{t_1-1}} \hat{\epsilon}_{t_1}, 
\end{equation}
where $\hat{\epsilon}_{t_1} = \epsilon_\theta(\mathbf{x}^1_{t_1},\mathbf{x}^1_c,\mathbf{x}^2_{t_1-\Delta t},\mathbf{x}^2_c,\mathbf{y})$ is the predicted noise, $\Delta t\geq0$ is a hyper-parameter to adjust the noise level of the two streams. We further follow 3DiM~\cite{3ddiffusion} to adopt an autoregressive sampling strategy, where the second stream loads previously generated views to guide the generation of the current view. In the training phase, two timesteps are randomly sampled, and in the sampling phase, we set $\Delta t$ to be a constant (except for the first view, where $\Delta t = 0$).


\begin{figure*}[!t]
    \centering
    \includegraphics[width=0.98\linewidth]{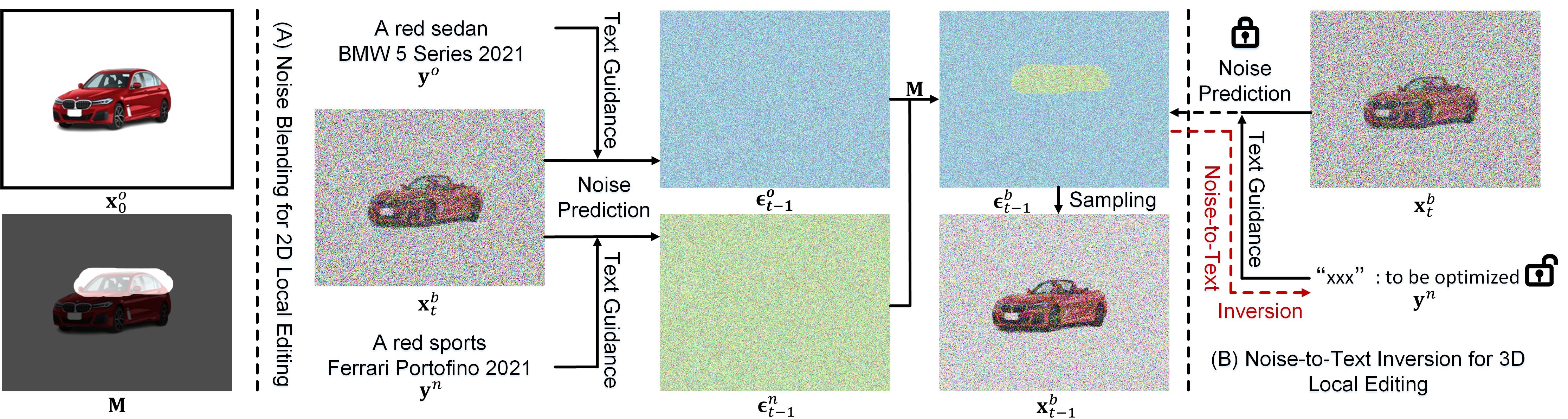}
    \caption{An illustration of 3D local editing. We propose to blend noises in each sampling step to achieve 2D local editing and conduct noise-to-text inversion to generate 3D manipulated images. The notations are explained in Sec.~\ref{sec:local}.}
    \label{fig:local}
\end{figure*}

\subsection{3D local editing}
\label{sec:local}
The model introduced above makes 3D-consistent visual content creation as simple as preparing a text. Moreover, to further achieve iterative refinement and fine-grained control, 3D local editing is required. 3D local editing takes a generated image (actually, the corresponding text), a single view mask that indicates the region to be manipulated, and a target text that guides the manipulation as inputs. The output would be manipulated images from different views. In the following, we first introduce our ``noise blending'' for 2D local editing. After that, we introduce a noise-to-text inversion process that enables our model to generate other views of the manipulated object. Figure~\ref{fig:local} shows an illustration to make the editing process clearer.


\noindent \textbf{Noise blending for editing a single view image.} We conduct spatial blending over the predicted noises to perform local editing for a given camera view. Formally, in each sampling step, we have:
\begin{equation}\label{eqn:local}
\begin{split}
\hat{\epsilon}^b_{t-1} & = (1-\mathbf{M})\odot \hat{\epsilon}^o_{t-1} + \mathbf{M}\odot  \hat{\epsilon}^n_{t-1}, \\
 \hat{\epsilon}^o_{t-1}  & =  \epsilon_\theta(\mathbf{x}^b_{t},\mathbf{y}^o),\\
\hat{\epsilon}^n_{t-1}  & =  \epsilon_\theta(\mathbf{x}^b_{t},\mathbf{y}^n),
\end{split}
\end{equation}
where $\hat{\epsilon}^b_{t-1}$ is the blended noise at timestep $t-1$, $\mathbf{M}$ is the user-provided mask that indicates the region to be manipulated, $\odot$ denotes the Hadamard product (\textit{i.e.,} element-wise product), $\hat{\epsilon}^o_{t-1}$ and $\hat{\epsilon}^n_{t-1}$ are the noise predicted based on the original $\mathbf{y}^o$ and new (target) text $\mathbf{y}^n$, respectively. $\epsilon_\theta$ is the trained diffusion model (we omit the coarse results and the other view for simplicity), $\mathbf{x}^b_{t}$ is the image that sampled by Equation~\ref{eqn:sampling} with the blended noise in the last timestep. 

\noindent \textbf{Noise-to-text inversion for generating 3D manipulated images.} To generate other views of the manipulated object, our insight is to invert the blended noise into the view-independent text space. Specifically, we \textit{fix the diffusion model and optimize a text embedding $\mathbf{y}^b$} to fit the blended noise in each sampling step:
\begin{equation}\label{eqn:loss_invert0}
L_{inv} = \Sigma_{t} \parallel \hat{\epsilon}^{b^{*}}_{t-1} -  {\epsilon_\theta}^{*}({\mathbf{x}^{b^{*}}_{t}},\mathbf{y}^b) \parallel,
\end{equation}
where $*$ indicates the fixed variables. In practice, we find that if the manipulated region is small, the optimization tends to be dominated by the original noise $\hat{\epsilon}^o_{t}$. Thus we extend the above Equation by adding a loss weight $\lambda$ to the manipulated region:
\begin{equation}\label{eqn:loss_invert}
\begin{aligned}
L_{inv} = & \Sigma_{t} \| (1-\mathbf{M}) \odot  ({\hat{\epsilon}^{o^*}_{t-1}}  -  {\epsilon_\theta}^*({\mathbf{x}^{b^{*}}_{t}},\mathbf{y}^b)) \\ & + \lambda \mathbf{M} \odot  ({\hat{\epsilon}^{n^*}_{t-1}}  -  {\epsilon_\theta}^*({\mathbf{x}^{b^{*}}_{t}},\mathbf{y}^b)) \| .
\end{aligned}
\end{equation}
Please refer to Equation~\ref{eqn:local} for the explanation of notations.

\begin{figure*}[!t]
    \centering
    \includegraphics[width=0.99\linewidth]{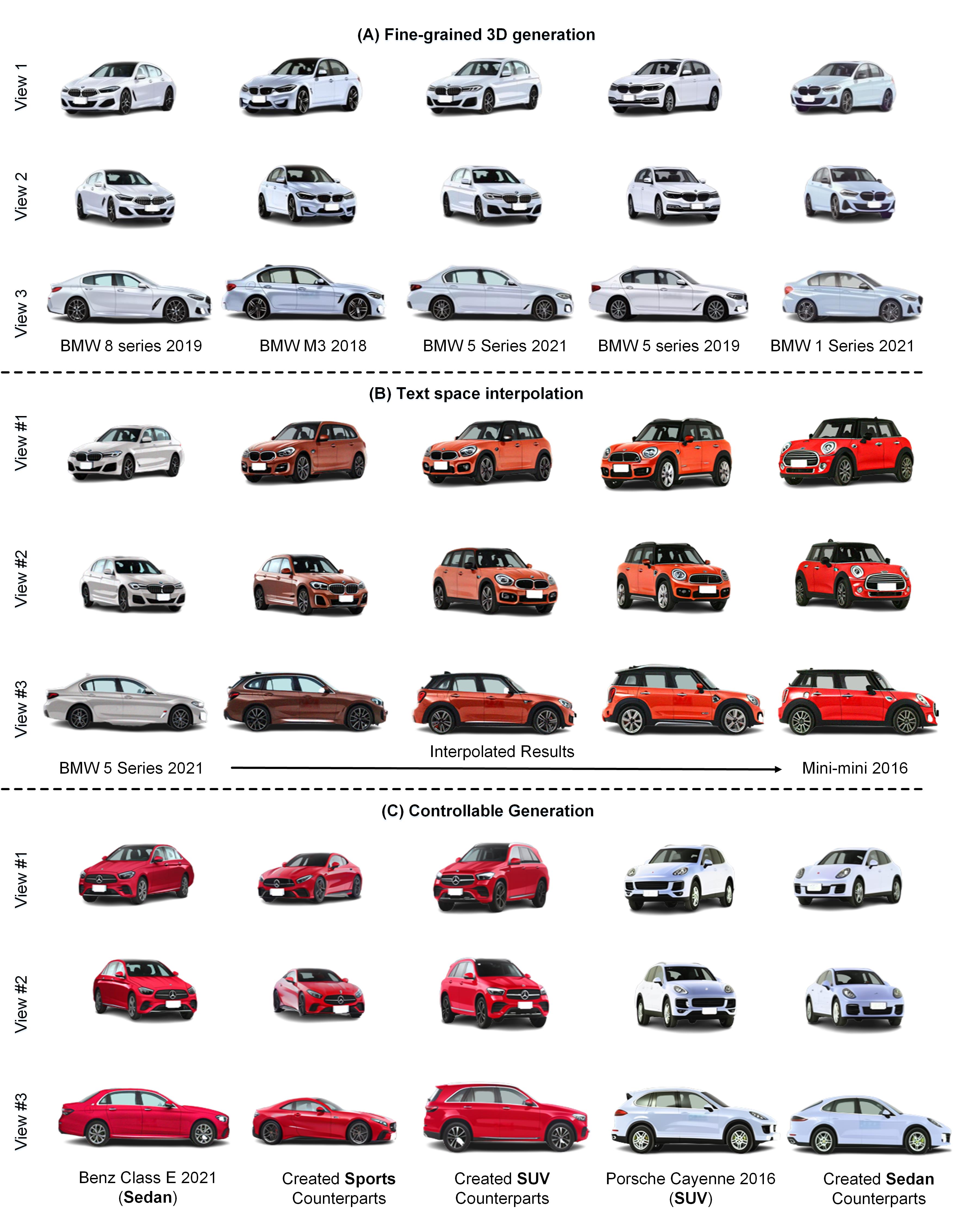}
    \caption{Our 3DDsigner can perform (A) fine-grained 3D generation, (B) semantic meaningful interpolation in the text embedding space, and (C) controllable generation that can change car types to create counterparts of real-world car models.}
    \label{fig:multi_tasks}
\end{figure*}

\noindent \textbf{Discussion and comparison.} We undertake a comprehensive comparative analysis and engage in in-depth discussions from three distinct perspectives: local editing, inversion, and alternative solutions.

\underline{\textit{Local editing.}} Our ``noise blending'' pipeline is similar to Blended-diffusion~\cite{avrahami2022blended}, which conducts spatially blending over the noisy image to edit a specific region of a given image. We propose to adapt the manipulation space from the noisy image space to the noise space, thus we can further invert the blended noise to a text embedding (as introduced in noise-to-text inversion) to achieve 3D local editing.

\underline{\textit{Inversion.}} 
The high-level insight of our noise-to-text inversion process is somehow similar to Textual inversion~\cite{gal2022image}, which learns a special text token with a pretrained diffusion model (fixed) to represent a specific concept (from 3-5 images). While an essential difference is that instead of following the fine-tuning paradigm, we learn the text embedding by fitting a whole DDIM sampling process, \textit{i.e.,} learning to fit every step that generates the manipulated object. Figure~\ref{fig:VisualizationLocalEditing} shows the effectiveness of our method.

\underline{\textit{An alternative solution.}} An alternative solution for 3D local editing would be generating a 2D manipulated image and further synthesising other views. 1) As discussed above, with a fixed diffusion model, image-to-text is much more difficult than noise-to-text inversion. 2) Fine-tuning diffusion models (Section~\ref{sec:novel}) can achieve one-shot novel view synthesis but not efficient. Since every time to fit an object, the diffusion model needs to be adjusted.
\vspace{-5mm}
\subsection{One-shot novel view synthesis}
\label{sec:novel}

In this study, we delve into a compelling question, namely, `How does our text-guided 3D generation model perform in the realm of novel view synthesis?' Specifically, one-shot novel view synthesis aims to synthesize new views of an object based on a single given image. Leveraging the prior object information learned by our model, we find that achieving one-shot novel view synthesis is easily attainable through fine-tuning on the provided image using Equation~\ref{eqn:loss}. To elaborate, we introduce noise to the given image and fine-tune our model to predict this added noise. During the fine-tuning process, we ensure that both views of our model align with the given view, and the timesteps are synchronized ($\Delta t = 0$). For detailed results and further discussions, please refer to Section~\ref{sec:exp_novel}.

It's crucial to highlight that only one given view is available for optimizing the two streams, and all parameters of our model are subject to training for the task of one-shot novel view synthesis.


%% file: latex/5experiment.tex
\section{Experiments}
\label{sec:exp}

\subsection{Experiment setup}

In this section, we present the experimental details of our approach. We outline the datasets, implementation details, baseline methods and evaluation metrics.

\begin{figure*}[!h]
  \centering
  \includegraphics[width=1.0\linewidth]{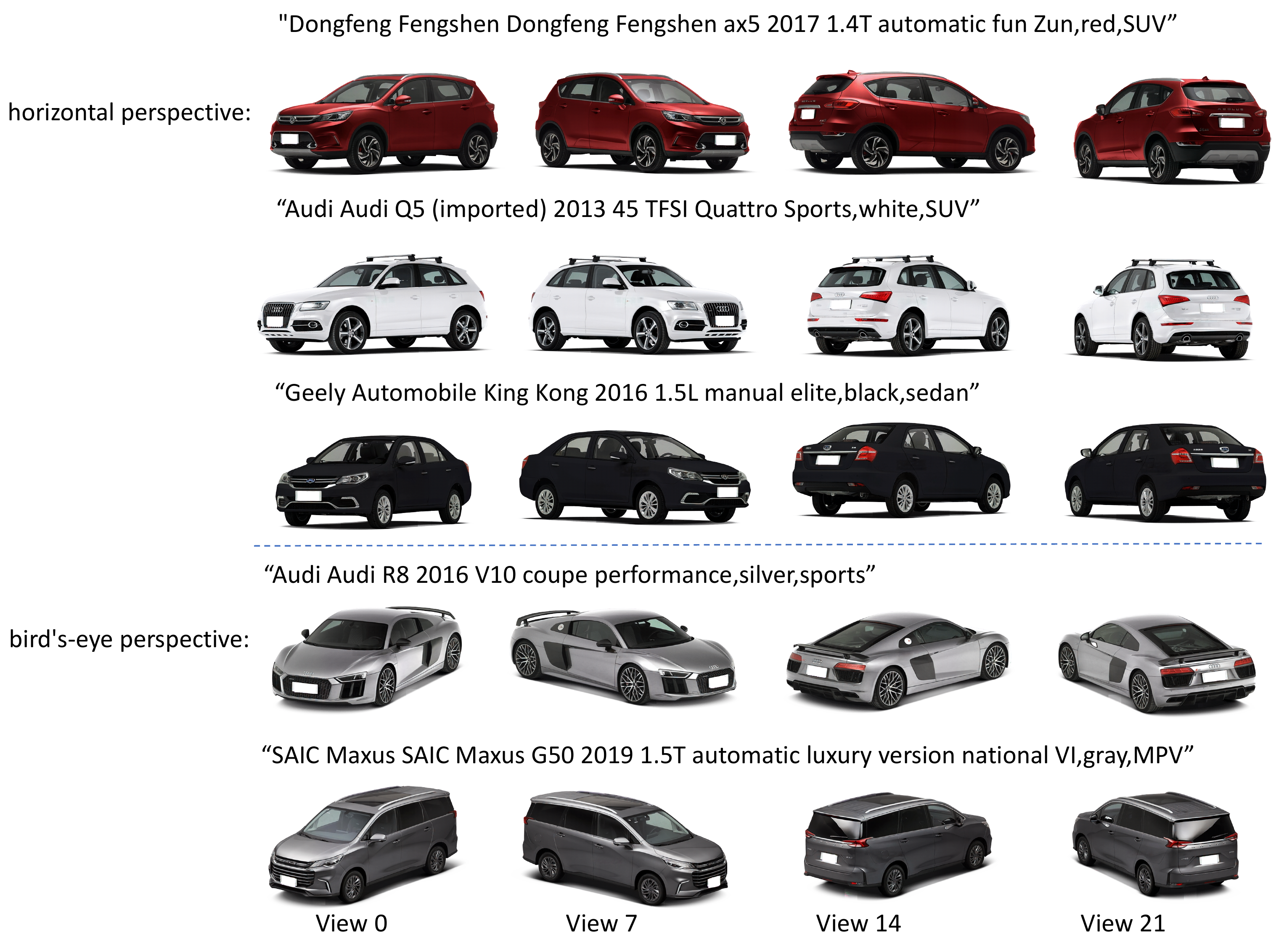}
   \caption{A set of examples consisting of image-text pairs sourced from our collected RealCars dataset.}
   \label{fig:show_samples}
\end{figure*}

\begin{table*}[]
\caption{Distribution of the various car types.}

\label{tab:carType}
\centering
\begin{tabular}{lcccccc}
\hline
Car Type      & SUV & Sedan  & Hatchback & Sports & MPV & Estate   \\ 

Num & 5854 & 3693 & 1040 & 636 & 443 & 115 \\ \hline
-  & Fastback & Hardtop convertible & Convertible & Pickup & Minibus & Crossover  \\
 - & 114 & 95 & 19 & 16 & 10 & 4 
\\ \hline

\end{tabular}
\end{table*}

\begin{table*}[]
\caption{Distribution of the various car colors.}

\label{tab:carColor}
\centering
\begin{tabular}{lccccccccc}
\hline
Car Color      & White & Blue  & Black & Red & Silver & Gray & Orange & Yellow & Green  \\

Num & 2737 & 2111 & 2029 & 1858 & 1307 & 909 & 615 & 549 & 161 
\\ \hline

\end{tabular}
\end{table*}

\begin{table*}[]
\caption{Distribution of the various car brands (only the top 50 brands with the highest quantity are shown).}

\label{tab:carBrand}
\centering

\setlength{\tabcolsep}{1mm}{
\begin{tabular}{lcccccccc}
\hline
Car Brand      & BMW & Audi  & Mercedes Benz & Volkswagen & Benz & Toyota & Porsche & BYD   \\

Num & 775 & 730 & 647 & 566 & 509 & 331 & 293 & 245 
\\ \hline

-   & Changan & Geely  & Honda & Chery & Volvo & Dongfeng  & Ford & Skoda    \\

-  & 233 & 222  & 221 & 218 & 218 & 217 & 210 & 189
\\ \hline

-  & Havel & GAC & Nissan  & Buick  & Mini & Aston Martin & Land Rover   & Hyundai  \\
-  & 177 & 172 & 171 & 166 & 162 & 156 & 155 & 151 \\ \hline

-  & Land Rover  & Lexus  & Cadillac & Jeep & Jaguar & Baojun  & SAIC & BAIC  \\
-  & 149 & 137 & 128 & 126 & 125 & 122 & 120 & 114 \\ \hline

-   & Maserati & Roewe & Lincoln & Infiniti & Tesla    & Mazda & Chevrolet & Qichen \\
- & 108 & 104 & 101 & 92 & 83 & 79 & 75 & 74 \\ \hline

\end{tabular}}
\end{table*}

\begin{figure*}[h]
    \centering
    \includegraphics[width=0.90\linewidth]{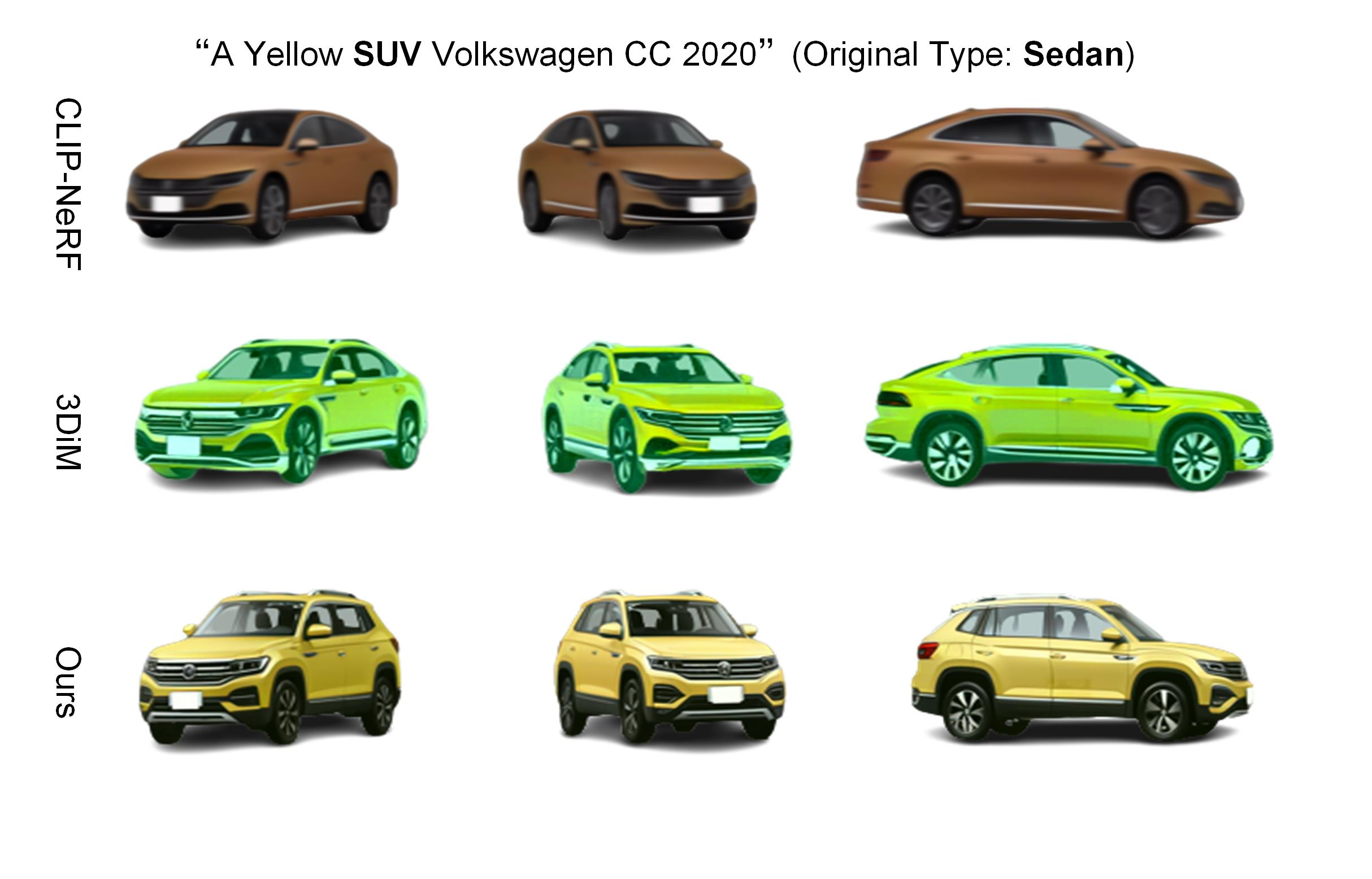}
    \vspace{-10mm}
    \caption{Visualization and comparison in terms of text-guided 3D generation. Our model can generate high-fidelity images that are more controllable than CLIP-NeRF~\cite{wang2022clip} and better 3D-consistent than extended 3DiM~\cite{3ddiffusion}.}
    
    \label{fig:cmp}
\end{figure*}

\begin{figure*}[!h]
    \centering
    \includegraphics[width=0.85\linewidth]{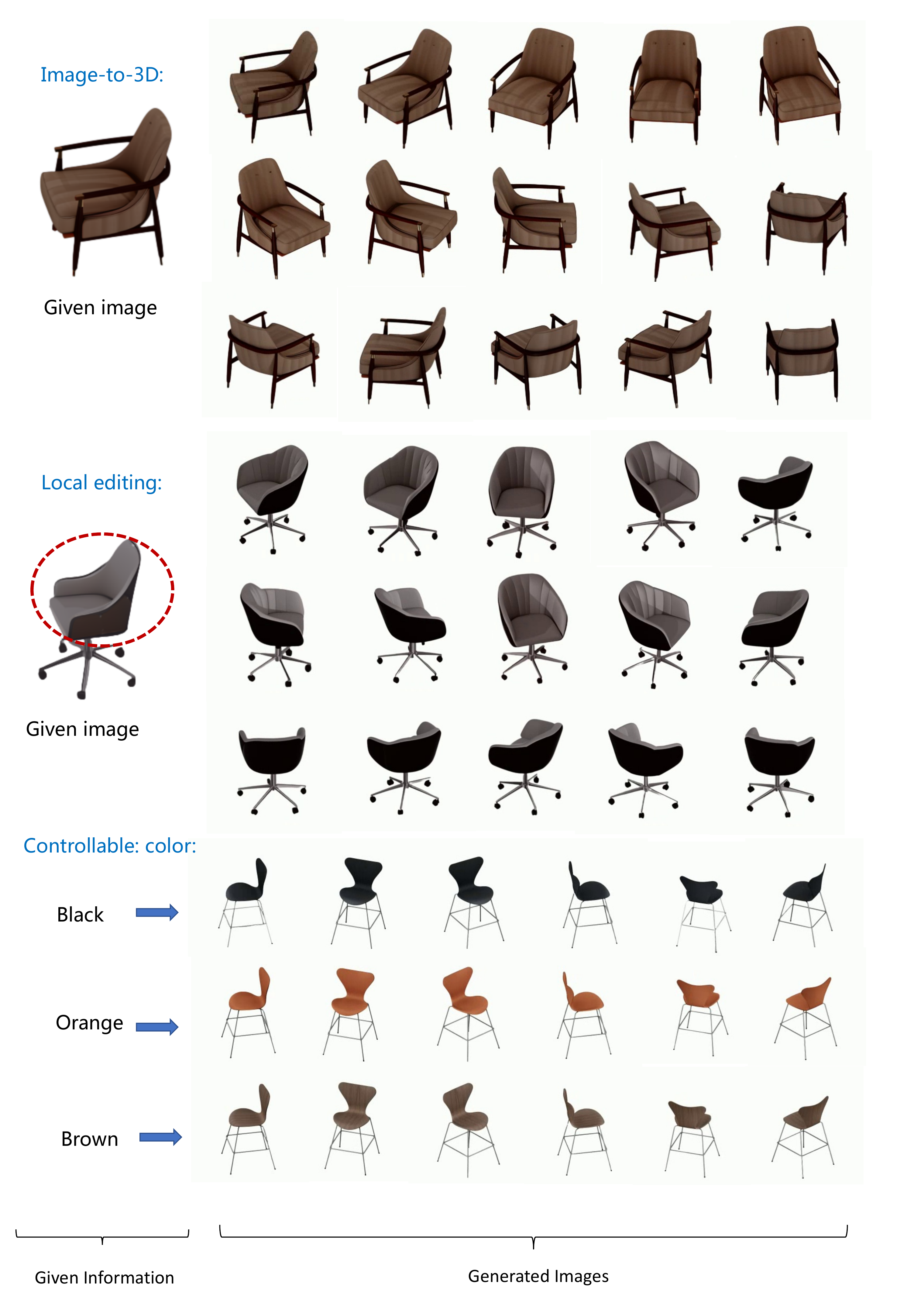}
    \caption{The generated results of chair category on 3D-FUTURE dataset. We  show the results of Image-to-3D, local loading, and color-controllable chair generation, respectively.}
    \label{fig:chairs}
\end{figure*}

\begin{figure*}[!h]
    \centering
    \includegraphics[width=0.90\linewidth]{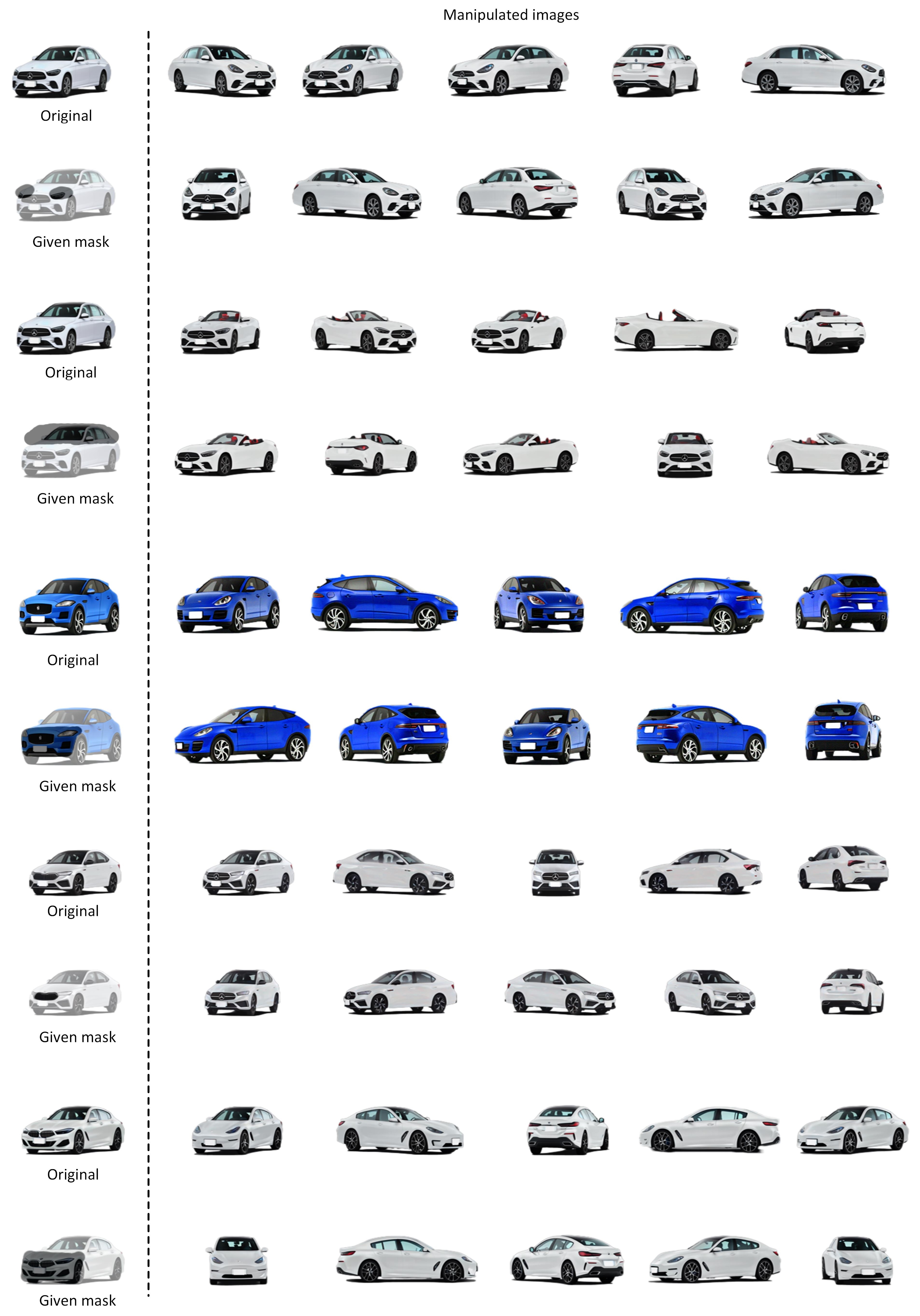}
    \caption{Visualization on text-guided 3D local editing. The left column shows the given original images and mask images, while the right column visualizes randomly selected 2D images from 10 different perspectives after local editing.}
    \label{fig:localediting}
\end{figure*}

\begin{figure*}[!h]
    \centering
    \includegraphics[width=0.90\linewidth]{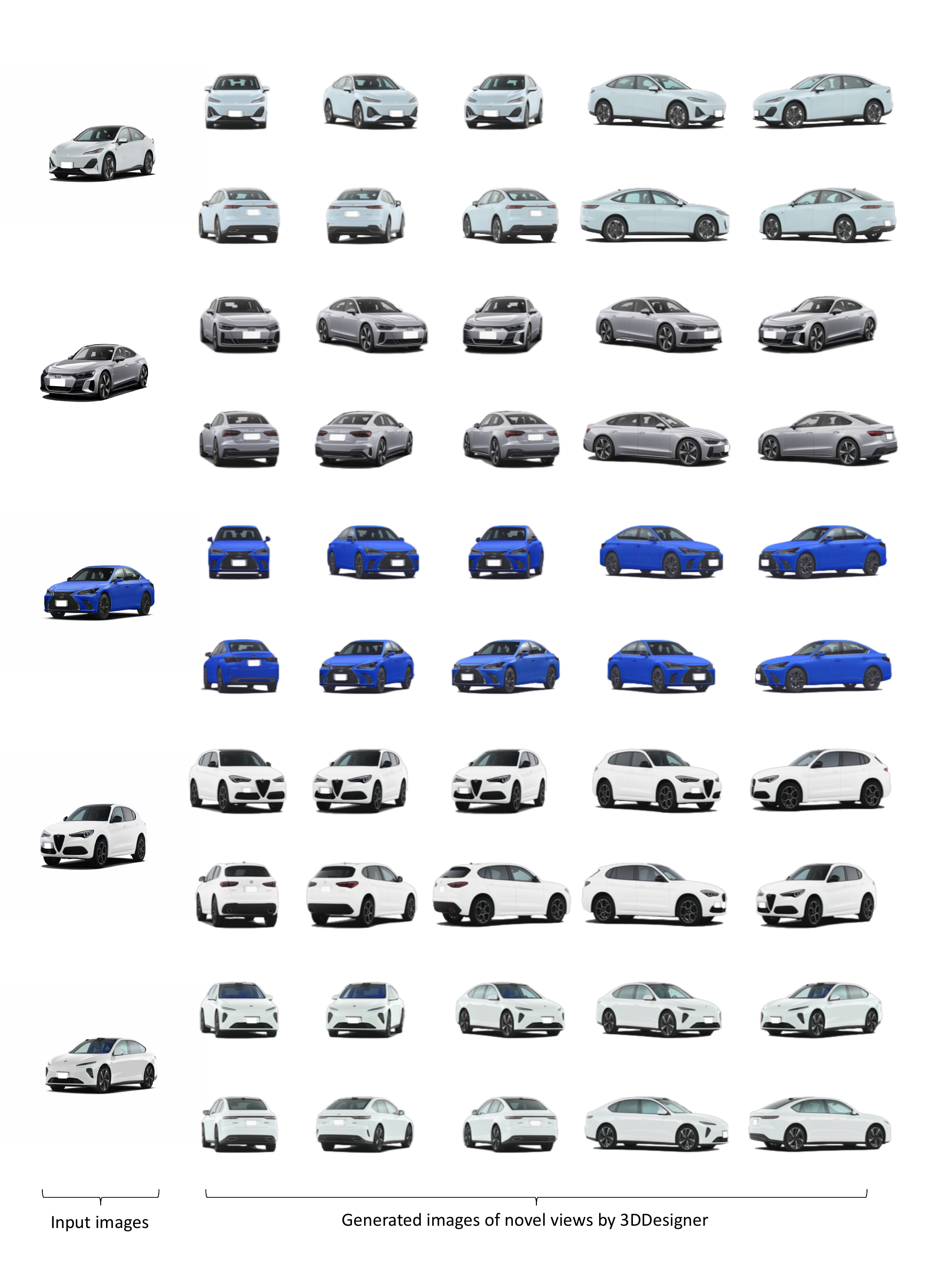}
    \caption{Visualization of One-Shot Novel View Synthesis. The left column displays the given original images, while the right column visualizes 2D images from 10 different perspectives generated through one-shot novel view synthesis.}
    \label{fig:novelview}
\end{figure*}

\begin{figure*}[h]
    \centering
    \includegraphics[width=1\linewidth]{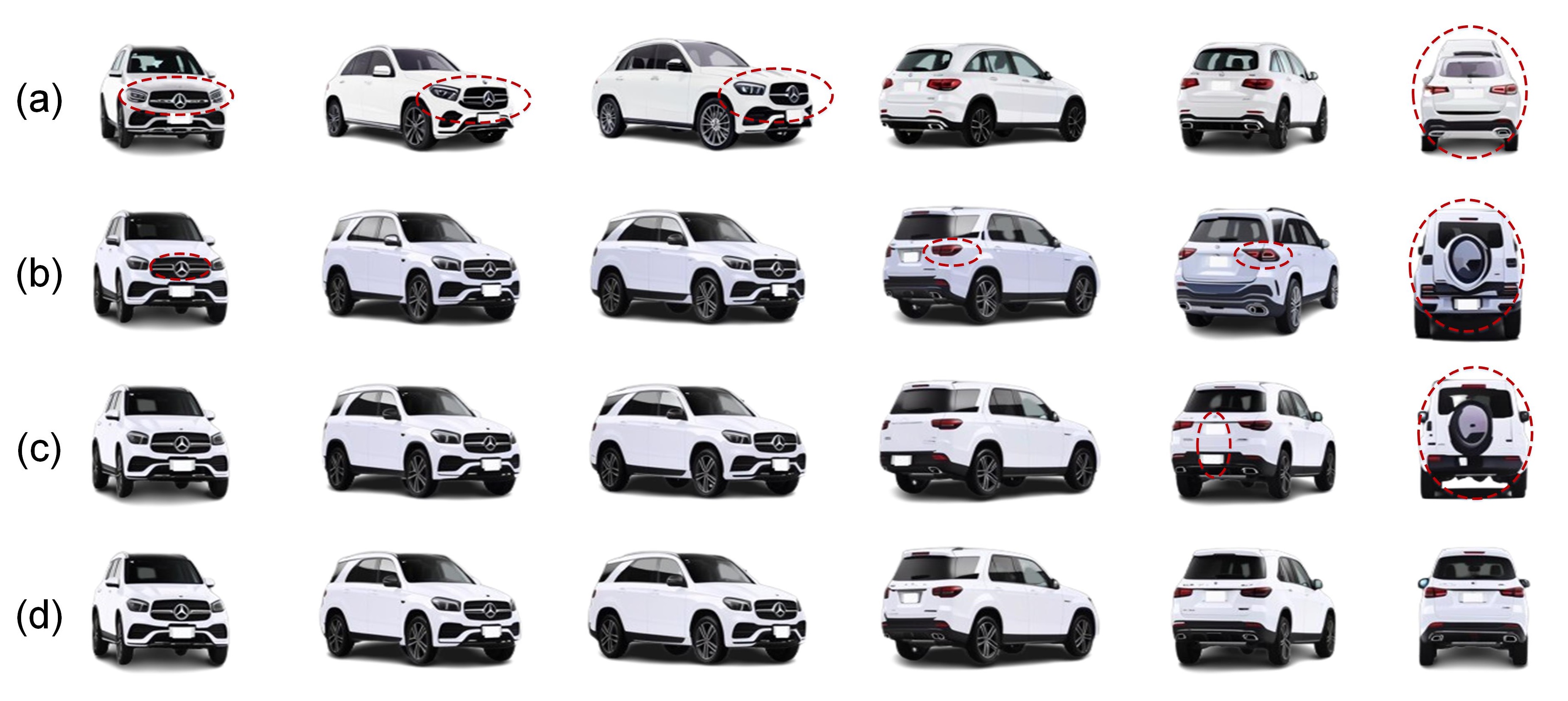}
    \caption{Ablation study on the two-stream asynchronous diffusion module: (a) single-stream diffusion, (b) synchronistic timesteps, (c) clean condition, and (d) ours. The red circles locate 3D-inconsistent parts. [Best viewed with zoom in]}
    \label{fig:ablation}
\end{figure*}

\begin{figure*}[t!]
    \centering
    \includegraphics[width=1\linewidth]{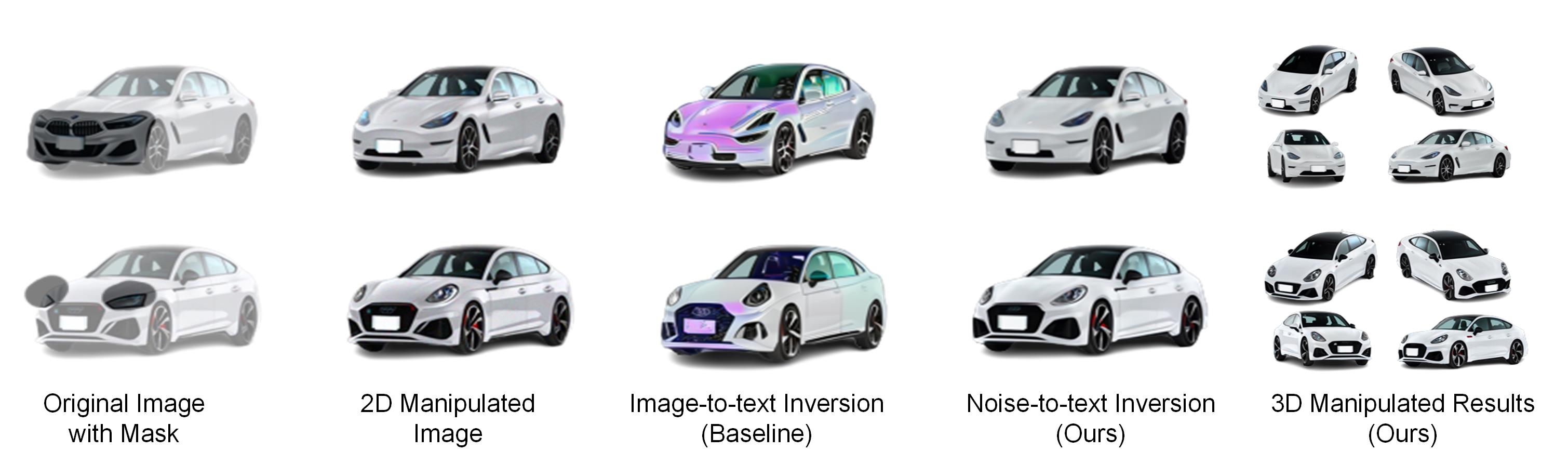}
    \caption{Visualization of 3D local editing. Noise-to-text inversion outperforms image-to-text inversion, and our model can generate 360$^\circ$ results given a single view mask.}
    \label{fig:VisualizationLocalEditing}
\end{figure*}

\begin{figure*}[t!]
    \centering
    \includegraphics[width=1\linewidth]{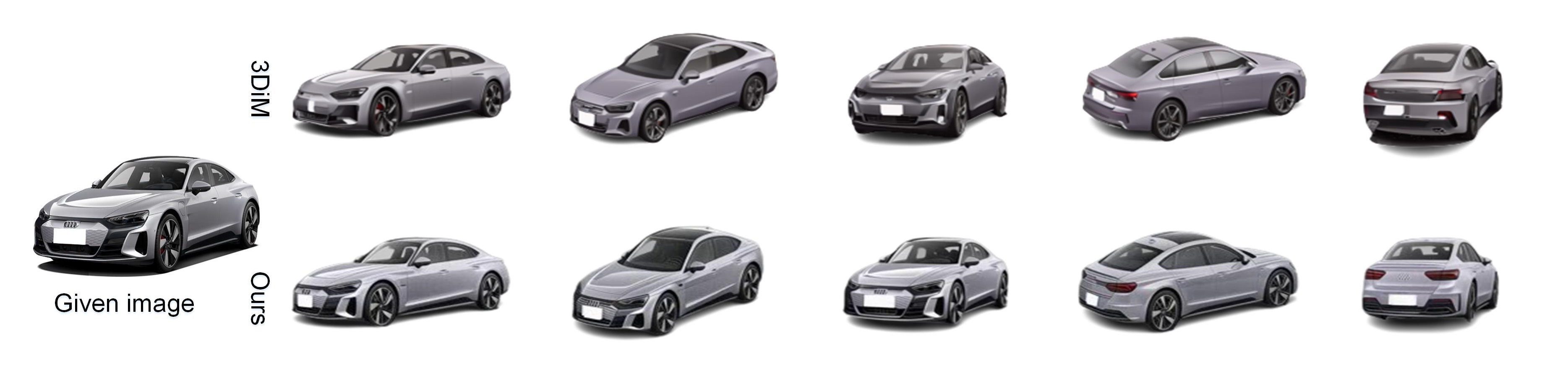}
    \caption{Visualization and comparison on one-shot novel view synthesis. Leveraging text guidance can significantly improve the performance of novel view synthesis.}
    \label{fig:novel}
\end{figure*}

\noindent \textbf{RealCars Dataset}. We collected a total of 493,530 images(including relevant textual information) from 11,696 cars, of which 4,755 cars have 60 views (including 30 views from a 360-degree horizontal perspective and 30 views from a 360-degree bird's-eye perspective, both at 12-degree intervals), and the remaining 6,941 cars have 30 views.  We train two classifiers on CompCars dataset to predict the color and type of each car to enrich text information. We present the distribution of the RealCars dataset by car type and color in Table \ref{tab:carType} and Table \ref{tab:carColor}, respectively. The collected text shows that our dataset contains approximately 202 brands of cars, and we display the top 40 brands in Table \ref{tab:carBrand}. We also randomly present a set of examples consisting of image-text pairs sourced from our dataset in Figure \ref{fig:show_samples}.
 We collect a 3D car dataset, which contains around 3k car models with corresponding texts (\textit{i.e.,} model names). Each model is rendered into 30 images from different camera views (12$^{\circ}$/360$^{\circ}$ per view) and augmented by around 4 different colors. Some models are rendered into 60 images with two different pitch angles. We also train a classification model on CompCars~\cite{yang2015large} dataset to predict the color and type of each car model. To this end, we can obtain a 3D car dataset with 400k images with texts, \textit{e.g.,} ``A red sedan BMW 5 series 2021''. We will make this dataset publicly released. For text-guided 3D generation, we create a testing set where the texts are novel compositions of the color, types, and model names. For novel view synthesis, the training and testing set contains different car models. 

 \noindent \textbf{3D-FUTURE dataset}. In addition to conducting experiments on the RealCars dataset that we proposed, we also conducted experiments on the chair category of the 3D-FUTURE dataset~\cite{fu20213d}.  3D-FUTURE is an extensive collection of industrial computer-aided design (CAD) models for furniture and interior finishes.
 Each shape is associated with multiple textures and materials, leading to the expansion of shape repositories. We opted for the relatively clean `chair' supercategory from this dataset for our experiments. To align more closely with our method of reconstructing 3D objects based on 2D images, we used the Blender tool to render textures of 3D models into 2D images from various viewpoints. We captured viewpoints at 12-degree intervals, resulting in a total of 30 different perspectives for each 3D model, which served as input data for our model.

 

\noindent \textbf{Implementation details}. We use Pytorch~\cite{paszke2019pytorch} as our codebase and the overall training of our model costs 14 days on 8 NVIDIA A100 GPUs. The model size is around 600Mb. During inference, we adopt DDIM~\cite{ddim} sampling with 100 steps. Note that the NeRF-based condition model only needs to run one time (instead of 100), thus the  inference time is around 2 times longer than traditional single-stream diffusion models. The loss weight $\lambda$ in Equation~\ref{eqn:loss_invert} is experimentally set to be 3.

\begin{table*}[t]
\caption{Quantitative Evaluation on text-guided 3D generation. 3DiM*: our extended version to support text-guided generation.}
\label{tab:compare}
\centering
\renewcommand\arraystretch{1.15} 
\setlength{\tabcolsep}{2.5mm}
\begin{tabular}{llcccc}
\hline
\multirow{2}{*}{Method}& \multirow{2}{*}{Components} & \multicolumn{2}{c}{Consistency} &  Quality   &   Controllability   \\ \cline{3-6} 
         & & PSNR $\uparrow$ & SSIM $\uparrow$ & FID $\downarrow$& Acc $\uparrow$(\%) \\ \hline
CLIP-NeRF~\cite{wang2022clip}    & NeRF          &   \textbf{35.10}   &   \textbf{0.95} &   62.97 & 10.90 \\
3DiM*~\cite{3ddiffusion}            & Diffusion           &   29.13   &  0.89    &  \underline{\emph{35.46}} & \underline{\emph{60.13}}   \\
\rowcolor{gray!20} Ours         & NeRF+Diffusion        &  \underline{\emph{30.84}}  &  \underline{\emph{0.93}}   & \textbf{18.77} & \textbf{71.33}  \\ \hline
\end{tabular}
\end{table*}

\noindent \textbf{Baselines.} Since there are few works exploring text-guided 3D object generation, we re-implement and compare two related models on our task to discuss the effectiveness and limits of NeRF and diffusion components. CLIP-NeRF~\cite{wang2022clip} leverages text features and NeRF to achieve multi-modal 3D object manipulation, we remove the CLIP branch and jointly learn a text embedding and a NeRF on our dataset. 3DiM~\cite{3ddiffusion} is a diffusion model designed for novel view synthesis, we extend their model to a text-guided version.

\noindent \textbf{Evaluation metric}. We evaluate our model from three dimensions, \textit{i.e.,} 3D consistency, image quality, and controllability. Specifically, 1) we follow 3DiM~\cite{3ddiffusion} to adopt a 3D consistency scoring, which can evaluate the 3D consistency of multi-view images without requiring ground truth images. 2) We adopt FID~\cite{heusel2017gans} to evaluate the image quality. Moreover, 3) we use a pre-trained classification model~\cite{yang2015large} to evaluate whether the types of the generated images are consistent with the given text.

\begin{table}[]
\caption{Ablation study (\textit{i.e.,} different $\Delta t$) of the asynchronous diffusion process on the PSNR and SSIM metrics. Although multi-stream U-Nets are also adopted in Make-A-Video~\cite{Make-A-Video} and 3DiM~\cite{3ddiffusion}, they use synchronistic timesteps and clean conditions, respectively.}
\label{tab:ablation}
\centering
\begin{tabular}{lccc}
\hline
Method                     & $\Delta t$  & PSNR$\uparrow$  & SSIM$\uparrow$    \\ \hline
Single-stream        &      --    &      28.56      &   0.87     \\
Synchronistic timesteps  &  0  & 30.62          & 0.92          \\
Clean condition         &   1000  & 30.68          & 0.91         \\
\rowcolor{gray!20}Ours  & 200 & \textbf{30.84} & \textbf{0.93} \\ \hline
\end{tabular}
\end{table}

\subsection{3D-consistent generation}
\noindent \textbf{Gallery.} Figure \ref{fig:multi_tasks} shows the visualization results of our 3DDesigner from three aspects, i.e., fine-grained 3D generation, semantic meaningful interpolation, and controllable generation. In particular, our model is capable of generating very fine-grained 3D objects (\textit{e.g.}, different series of BMW as shown in Figure \ref{fig:multi_tasks} (A)), showing the large capacity of our model and making it more practical for real-world scenarios. Moreover, we also find that our model can produce semantically meaningful interpolations as shown in Figure \ref{fig:multi_tasks} (B), which opens a convenient gateway to create novel car models. Last but not least, our 3DDesigner shows great performance in terms of controllability, \textit{e.g.}, we can effectively change the type of cars from sedan to sports and SUV as shown in Figure \ref{fig:multi_tasks} (C). In addition to presenting the generated results on the RealCars dataset, we also demonstrate the generated outcomes on the 3D-FUTURE dataset in Figure \ref{fig:chairs}. We showcase our model's consistency in view generation, local editing capabilities, and attribute modifications (e.g., changing the color of chairs) 

\noindent \textbf{Comparison.} To the best of our knowledge, we are the first to extend text-guided diffusion models to achieve photorealistic 3D-consistent generation. Thus there is no available previous work that can be compared directly. To discuss the effectiveness and limits of NeRF and diffusion components, we re-implement and extend CLIP-NeRF~\cite{wang2022clip} and 3DiM~\cite{3ddiffusion} to our task. It can be observed from Table~\ref{tab:compare} that CLIP-NeRF~\cite{wang2022clip} performs best on 3D consistency, however, lacking diffusion model makes it fail to achieve promising image quality and controllability. Adding GAN loss may improve the visual quality, while the controllability is hard to be improved in their design. Note that 3DiM originally cannot perform text-guided generation, thus we integrate 3DiM with text guidance~\cite{Glide}. Compared to our model, the extended 3DiM: 1) lacks of the NeRF-based condition module and 2) adopts a different two-stream diffusion process. The results show that our model can significantly improve 3D consistency, image quality, and controllability. Moreover, visualization results in Figure~\ref{fig:cmp} also show the superior performance of our model.

\noindent \textbf{Ablation study.} The comparison to extended 3DiM~\cite{3ddiffusion} in Table~\ref{tab:compare} has shown the effectiveness of our NeRF-based condition model. Here we take a closer look at our proposed two-stream asynchronous diffusion module, where a cleaner condition stream is designed to guide a generation stream. The results in Table~\ref{tab:ablation} show that 1) without the two-stream architecture to learn cross-view correspondence, the 3D consistency drops seriously; 2) without asynchronous timesteps, the conditioned stream is as noisy as the generation stream, leading to poor guidance; 3) conditioned on a clean image cannot well leverage the guidance due to the large gap of noise level. Figure~\ref{fig:ablation} shows that our proposed two-stream asynchronous diffusion can generate more consistent results.

\vspace{-5mm}
\subsection{3D local editing}
We propose to extend the 2D editing method (\textit{i.e.,} noisy image blending~\cite{avrahami2022blended}) to noise blending, thus the blended noise can further be inverted into a view-independent text embedding for 3D local editing. The motivation of such a design is that inverting noises in a specific DDIM sampling process is much easier than inverting an image. It can be observed from the third column of Figure~\ref{fig:VisualizationLocalEditing} that obvious artifacts appear in image-to-text inversion results. The last two columns of Figure~\ref{fig:VisualizationLocalEditing} show that our proposed method can generate high-fidelity 360$^\circ$ manipulated images with a single view mask. We provide additional results of 3D local editing in Figure \ref{fig:localediting}. A profusion of visualizations serves to underscore the broad applicability of our methodology.

\begin{table}[]
\caption{Quantitative Evaluation of One-Shot Novel View Synthesis, comparing our approach with concurrent work based on diffusion models using PSNR and SSIM metrics.}
\label{tab:novel}
\centering
\newcommand{\tabincell}[2]{\begin{tabular}{@{}#1@{}}#2\end{tabular}}
\renewcommand\arraystretch{1.1} 
\setlength{\tabcolsep}{1.0mm}{
\begin{tabular}{lcc}
\hline
Method    & PSNR $\uparrow$  &   SSIM $\uparrow$   \\ \hline
3DiM~\cite{3ddiffusion}                  &    32.53       &   0.93    \\
\rowcolor{gray!20}Ours & \textbf{33.14} & \textbf{0.94} \\ \hline
\end{tabular}}
\end{table}

\subsection{One-shot novel view synthesis}
\label{sec:exp_novel}
Inspired by recent advances in fine-tuning text-guided diffusion models for image inversion~\cite{gal2022image,kawar2022imagic}, we extend our text-guided 3D generation model to perform novel view synthesis. Given an object that the model has nerve seen, we fine-tune our model by adding noise to a single-view image of the object and conduct denoising. After that, we can obtain novel view images by changing the input camera pose. Table~\ref{tab:novel} and Figure~\ref{fig:novel} shows the qualitative and quantitative results, respectively. Figure~\ref{fig:novel} illustrates the comparative advantages of our model over 3DiM~\cite{3ddiffusion}. It can be observed that our model can synthesis more realistic and consistent novel view images. To comprehensively showcase our model's generation capabilities in one-shot novel view synthesis, we provide additional visual results in Figure~\ref{fig:novelview}.

%% file: latex/6conclusion.tex
\section{Conclusion}
\label{sec:conclusion}

In this paper, we introduced 3DDesigner, a text-guided diffusion model which is capable of generating and manipulating 3D objects. To our best knowledge, the proposed 3DDesigner is the first text-guided generative model that can perform 3D-consistent generation, 3D local editing, and one-shot novel view synthesis together. Benefitting from the proposed NeRF-based Condition Module, Two-stream Asynchronous Diffusion Module, and novel diffusion training, sampling\&blending strategies, our 3DDesigner outperforms existing methods and achieves highly realistic, detailed, and controlled 3D object generation.

\section{Data availability statement}
\label{sec:conclusion}

The data that support the findings of this study are openly available, including ``3D-FUTUR" and ``RealCars," which are included in this published article~\cite{fu20213d} and the link: \url{https://pan.baidu.com/s/1d6zC3oN2pq2frd3bbnYt-Q?pwd=veeb}, respectively.